\documentclass{article}

\usepackage[final]{neurips_2024}

\usepackage[utf8]{inputenc} 
\usepackage[T1]{fontenc}    
\usepackage{hyperref}       
\usepackage{url}            
\usepackage{booktabs}       
\usepackage{amsfonts}       
\usepackage{nicefrac}       
\usepackage{microtype}      
\usepackage{xcolor}         

\usepackage{subcaption}
\usepackage{graphicx}

\usepackage{footmisc}
\usepackage{booktabs}
\usepackage{tablefootnote}
\usepackage{amsmath}
\usepackage{caption}
\usepackage{placeins}
\usepackage{pgfplots}
\usepackage{pgfplotstable}
\usepackage{tikz}
\usepackage{adjustbox}
\usepackage{amsmath}
\usepackage{float}
\usepackage{footnote}
\usepackage{etoolbox}
\usepackage{array, booktabs, makecell}
\usepackage{xspace}
\usepackage{tabularx}
\usepackage{multirow}
\usepackage{adjustbox}
\usepackage{multicol}
\usepackage{soul}
\usepackage{txfonts}
\usepackage{wrapfig}

\usepackage[capitalise]{cleveref}
\crefname{section}{Sec.}{Sec.}
\Crefname{section}{Section}{Sections}
\crefname{figure}{Fig.}{Figs.}
\Crefname{figure}{Figure}{Figures}
\crefname{appendix}{App.}{Apps.}
\Crefname{appendix}{Appendix}{Appendices}
\crefname{table}{Tab.}{Tabs.}
\Crefname{table}{Table}{Tables}
\crefname{footnote}{Fn.}{Fns.}
\Crefname{footnote}{Fn.}{Fns}

\newcommand{\com}[1]{}
\newcommand{\ourmethod}{DISCO\xspace}

\newcommand{\starcodertarget}{\texttt{Starcoder-15B}\xspace}
\newcommand{\starcoderdraft}{\texttt{Starcoder-168M}\xspace}
\newcommand{\vicunatarget}{\texttt{Vicuna-13B}\xspace}
\newcommand{\vicunadraft}{\texttt{Vicuna-68M}\xspace}

\newcommand{\vicuna}
[0]{\texttt{Vicuna}\xspace}

\newcommand{\starcoder}[0]{\texttt{Starcoder}\xspace}

\title{Dynamic Speculation Lookahead Accelerates\\Speculative Decoding of Large Language Models}

\author{\textbf{Jonathan Mamou}$^\varspadesuit$ \quad\textbf{Oren Pereg}$^{\varspadesuit}$ \quad
\textbf{Daniel Korat}$^\varspadesuit$ \quad \\ \textbf{Moshe Berchansky}$^\varspadesuit$ \quad \textbf{Nadav Timor}$^\diamondsuit$ \quad 
\textbf{Moshe Wasserblat}$^\varspadesuit$ \quad \textbf{Roy Schwartz}$^\varclubsuit$ \\
$^\varspadesuit$ Intel Labs, Israel\\  
$^\diamondsuit$ Weizmann Institute of Science, Israel\\
$^\varclubsuit$School of Computer Science \& Engineering, Hebrew University of Jerusalem\\
\small{
   \textbf{Correspondence:} \href{mailto:jonathan.mamou@intel.com}{jonathan.mamou@intel.com}
 }
}

\pgfplotsset{compat=1.18}
\begin{document}
\maketitle
\begin{abstract}
    Speculative decoding is commonly used for reducing the inference latency of large language models.
    Its effectiveness depends highly on the \textit{speculation~lookahead}~(\textit{SL})---the number of tokens generated by the draft model at each iteration.
    In this work we show that the common practice of using the same SL for all iterations (\textit{static SL}) is suboptimal.
    We introduce {\ourmethod} (\textbf{D}ynam\textbf{I}c \textbf{S}pe\textbf{C}ulation lookahead \textbf{O}ptimization), a novel method for dynamically selecting the SL.
    Our experiments with four datasets show that \ourmethod reaches an average speedup of 10\% compared to the best static SL baseline, while generating the exact same text.
    
\end{abstract}

\section{Introduction}

Large language models (LLMs) generate tokens autoregressively~\citep{radford2019language,brown2020language}, which often leads to slow generation.
Speculative decoding algorithms~\cite{Leviathan2023, SpecInfer_Miao2023, Chen2023_deep_mind} reduces the inference latency by splitting the inference of a given model into two steps. First, a fast \textit{draft} model generates tokens autoregressively. Then a more accurate~(\textit{target}) model validates all generated draft tokens simultaneously. See~\cref{spec_dec_fig} for an example. 
%
The effectiveness of speculative decoding depends on the \textit{speculation~lookahead}~(\textit{SL})---the number of tokens generated by the draft model at each iteration.
An SL too small leads to too many target forwards steps; 
SL values too large add redundant draft forward passes.
Yet, existing speculative decoding approaches use a \textit{static SL}---an SL that remains constant across all iterations~\citep{Leviathan2023, SpecInfer_Miao2023, Chen2023_deep_mind, Medusa_2024, Eagle_2024}. This work starts by defining an oracle SL---a method that assigns each iteration its optimal SL. 
We observe that the optimal SL shows a high variance across iterations~(\cref{fig:oracle:ex}).
We then use this oracle to estimate an upper bound of the expected speedup (compared to using static SLs), showing a potential gain of up to 39\% speedup.
We then propose \textit{\ourmethod}, a novel method for selecting the SL before each iteration.
\ourmethod estimates the likelihood of the next draft token being accepted by the target model, and halts the draft model if this likelihood is too small.
We evaluate \ourmethod across various tasks: code generation, text summarization, and instruction following.
Our results show an average speedup of 10\% compared to optimal static SL and 31\% compared to a previously known heuristic for controlling SLs~\cite{gante2023assisted}, all without modifying the output text (\cref{sec:app_adtnl_results}).
\ourmethod also transfers well across tasks from the same category: training it on one task and using it on another leads to similar speedups.

\section{Background: Speculative Decoding}
Speculative decoding expedites LLM generation while ensuring no accuracy loss by dividing it into two stages.
In the first stage, a fast but less accurate \textit{draft} model $M_D$ autoregressively generates a sequence of tokens. In the second stage, a large but more accurate \textit{target} model $M_T$ conducts parallelized verification over the generated draft tokens. This process allows the model to potentially produce multiple tokens per target forward pass.

\begin{wrapfigure}{r}{0.4\textwidth}
    \includegraphics [width=37mm]{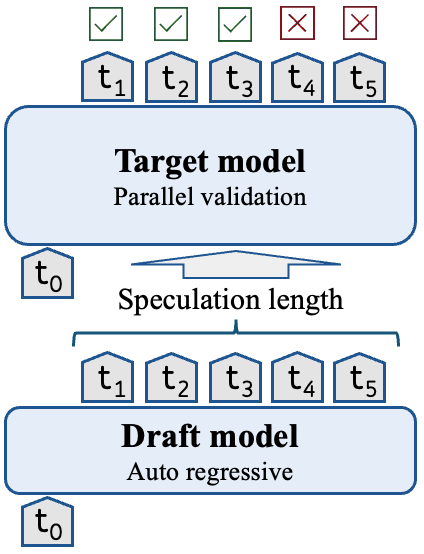}
    \centering
    \caption{An illustration of a single speculative decoding iteration with Speculation Lookahead~(SL)~=~5. Given a prompt $t_0$, a draft model  autoregressively generates 5 tokens $t_1, \ldots, t_5$. The target model validates them all in parallel and accepts only $t_1, t_2, t_3$. As $t_4$ and $t_5$ are rejected, the SL is suboptimal~(too large).}
    \label{spec_dec_fig}
    \vspace{-5mm}
\end{wrapfigure} 

\paragraph{Speculation lookahead (SL)}\label{sec:spec_len}
The effectiveness of speculative decoding in accelerating the token generation process relies heavily on the SL parameter, which determines how many tokens are generated by the draft model before each validation step. 
The effect of the SL on the overall speedup is subject to a tradeoff; higher SLs potentially reduce the number of target model validations, but also increase the number of redundant draft generations~(\cref{spec_dec_fig}), and vice-versa. 
The majority of speculative decoding approaches use a static SL---the same number of draft tokens are generated per speculative iteration.\footnote{A notable exception is \citet{gante2023assisted}, who applies a heuristic for dynamic SL adaptation by modifying the SL based on the acceptance rate of previous iterations.} 
\citet{Chen2023_deep_mind} explored various static SLs, across different target-draft model pairs and tasks, and empirically showed that as the SL rises, the overall speedup increases until reaching a certain threshold, beyond which it either levels off or even regresses. 
To study the effect of the SL, \citet{Leviathan2023} defined the improvement factor (IF) as the expected token generation speedup: 

\begin{equation}\label{eq:IF}
IF = {\frac{1-\alpha^{\gamma+1}}{(1-\alpha)({\gamma}c+1)}}
\end{equation}

where $\alpha$ denotes the acceptance rate, indicating the expected probability of a draft token to be accepted by the target model; $c$ represents the cost coefficient, indicating the ratio between the walltime of a forward pass run of the draft model $M_{D}$ and the wall time of a forward pass run of the target model $M_{T}$; and $\gamma$ represents the static SL value.\footnote{IF computation assumes enough compute resources for increased concurrency as $\gamma$ rises.} 
While both $\alpha$ and $c$ are important to the selection of the target-draft model pair, finding the optimal $\gamma$ is fundamental to the effectiveness of the system.

\section{Dynamic Speculation Lookahead}
The IF function~(\cref{eq:IF}) is based on the simplifying assumption that the probability of accepting draft tokens by the target model is i.i.d. Nevertheless, in practical scenarios, different tokens may have varying levels of predictability, which challenges this i.i.d.~assumption, and suggests that using a static SL might be suboptimal. Below we consider an oracle experiment, which applies the optimal dynamic $\gamma$ value at each iteration. We then propose a method for dynamically setting $\gamma$, showing that it strongly outperforms a static selection method for any choice of a static~SL.


\begin{figure}[t]
    \begin{minipage}{0.49\textwidth} 
        \vspace{0pt}
        \centering
        \includegraphics[width=0.8\textwidth]{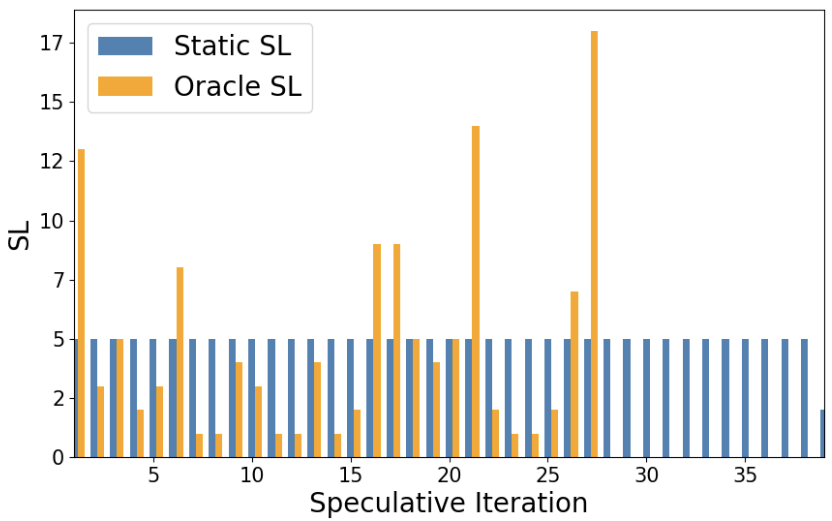}
        \caption{Oracle and static SL values for different speculative iterations on one MBPP example. For static SL, we run 38 target forward passes and 192 draft forward passes, while for oracle SL, we only run 27 target forward passes and 129 draft forward passes. We observe a high variance of oracle SL values.}
        \label{fig:oracle:ex}
    \end{minipage}
    \hfill
    \begin{minipage}{0.49\textwidth}
        \vspace{-35pt}
        \centering
        \includegraphics[width=0.845\textwidth]{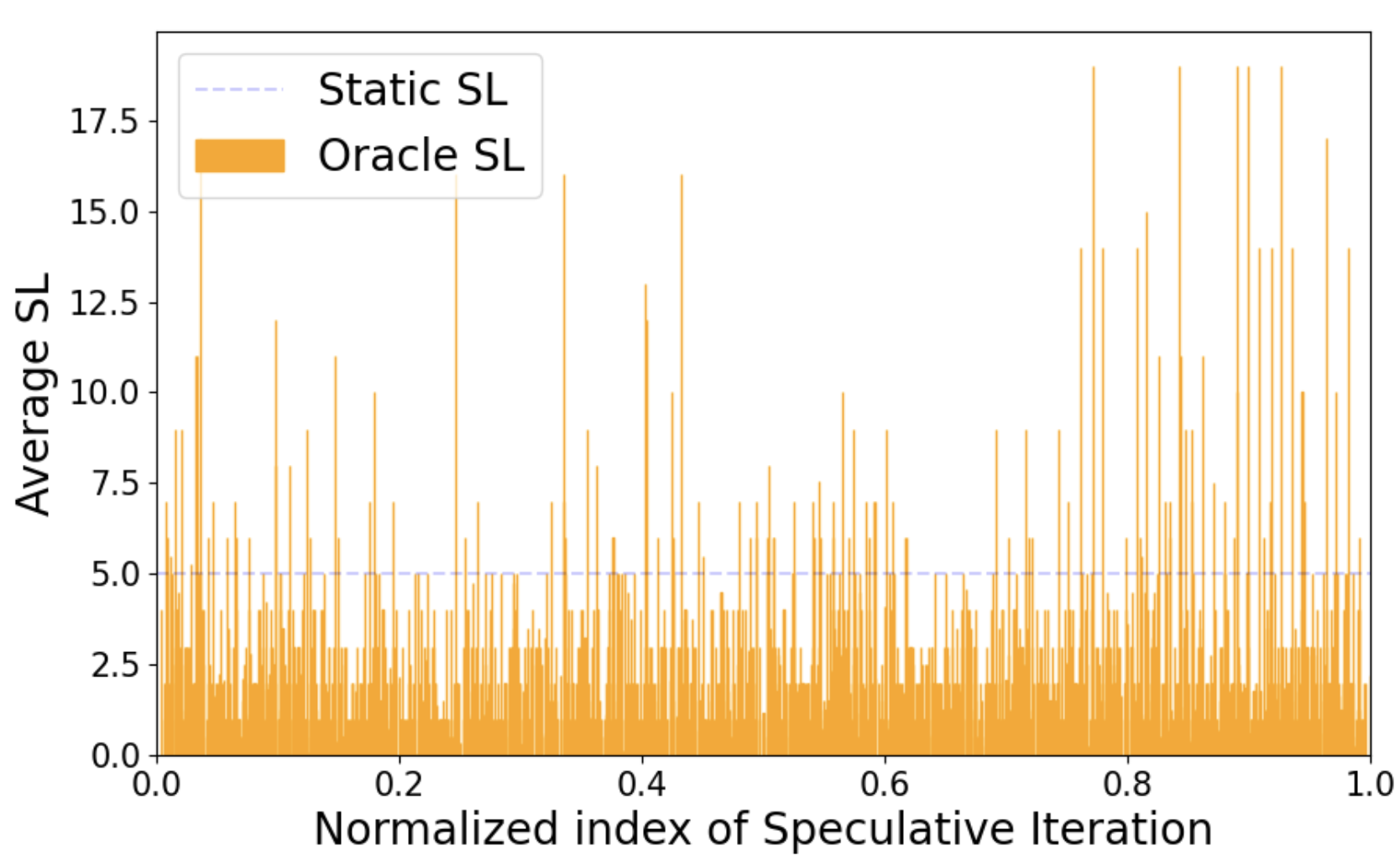}
        \caption{The average oracle SL over the normalized index of the speculative iterations for the Alpaca dataset. We observe a high variance of oracle SL values.}
        \label{fig:oracle:alpaca}
    \end{minipage}
\end{figure}

\paragraph{Finding the optimal SL per iteration} We start by employing an oracle for detecting the optimal value of SL ($\gamma$) for each speculative iteration. The oracle uses the draft model to autoregressively generate tokens until a mismatch occurs between the predicted tokens of the draft and target models. This process is repeated for each speculative iteration, ultimately returning the optimal (maximum) number of accepted draft tokens per iteration. 
The mismatch between the tokens is determined by using the rejection sampling algorithm introduced by \citet{Leviathan2023} with zero temperature.
This oracle  fulfills the speculative decoding potential: generating the maximal number of valid draft tokens at each iteration, and making a minimal number of calls to both draft and target model. 
\Cref{fig:oracle:ex} shows the oracle SL values across the speculative iterations for one MBPP example. Compared to the static SL, we observe a lower number of both  draft and target forward passes.
\Cref{fig:oracle:alpaca} shows the average oracle SL over the speculative iterations for the Alpaca dataset~\cite{alpaca2023}. Both figures show a high variance of oracle SL values, implying that astatic SL is likely to be suboptimal. See \cref{sec:appendix_oracle} for further analysis.


 

\paragraph{DynamIc SpeCulation lookahead Optimization}
We introduce \ourmethod, a simple method for dynamically setting the SL value at each iteration. 
To estimate the correct SL value at each step, we employ a simple classifier as follows. Immediately after generating any draft token, our classifier decides whether the draft model $M_{D}$ should proceed and generate the next token or switch to the target model $M_{T}$ for verification. The classifier takes as inputs the probability vector of the draft model ($y^{D}_i$) and the token position ($i$), generating a confidence score ($C_i$) used for the decision-making as follows:
\begin{equation} 
    C_i=FFN(Concat(Top_k(y^{D}_i),Ent(y^{D}_i),i))
\end{equation}

where $Top_k()$ selects the top $k$ values and $Ent()$ is the entropy function.\footnote{We use $k=10$ in all experiments.} 
At inference time, $C_i$ is compared against a predetermined threshold $\tau$ to decide whether the draft model should continue to generate the next token or turn to the target model for validation. In addition, we limit the number of draft generated tokens to $\text{SL}_{\max}$.\footnote{$\text{SL}_{\max}$ enables optimized execution with LLMs using fixed tensor shapes.\label{SL_max}} Note that our method adapts the rejection sampling algorithm which preserves the distribution of $M_{T}$, thus ensuring no quality degradation.



\section{Experiments}
\paragraph{Datasets and Models}
We evaluate our method on four datasets spanning three tasks: code generation using MBPP~\cite{MBPP_austin2021} and HumanEval~\cite{HumanEval2021}; text summarization using CNN-DailyMail~\cite{CNN_DM_2016}; and instruction-following using Alpaca. We use the training sets for training the SL classifier and the validation sets for setting the threshold $\tau$ and the $\text{SL}_{\max}$ hyperparameter. For HumanEval, which has no training and validation sets, we evaluate transfer learning from MBPP.
For code generation tasks, we use the \starcoder model family~\cite{starcoder_2023}---15B for target and 168M for draft.
For the other tasks, we use \vicuna models---13B as target~\cite{vicuna_2023} and 68M as draft~\cite{tiny_vicuna_2024}. 
See~\cref{sec:prompts,sec:models} for more details. 

\paragraph{SL classifier training}
To train the classifier we extract features from the training sets of our datasets MBPP, CNN-DM, and Alpaca. For minimal overhead, we use a shallow 2-layer FFN classifier and train it based on the extracted features to predict the agreement between the draft and target models. The training  employs cross-entropy loss with Total Variance as distance measure: $TV(y^D_i,y^T_i)$, where $y^D_i$ and $y^T_i$ represent the vocabulary distribution of $M_{D}$ and $M_{T}$ respectively at the $i^{th}$ token position. 
We evaluate the quality of the classifier by measuring its F1 score on the validation set. The F1 results obtained on the datasets are relatively high; for instance, 95\% on MBPP, compared to 85\% using the optimal static SL. See~\cref{sec:app_cls,sec:app_impl} for more details.

\paragraph{Baselines and Results}

We compare the LLM inference latency of \ourmethod to both static SL (\textit{static SL-5}) and dynamic heuristic SL setups (\textit{dynHeur SL}; \citealp{gante2023assisted}). 

\begin{wraptable}{r}{0.53\textwidth} 
	\centering
	\small
    \begin{tabularx}{1.0\linewidth}{p{1.5cm}lcc} 
    \toprule[\heavyrulewidth]\toprule[\heavyrulewidth]

    \textbf{Benchmark} & \textbf{Method} & \multicolumn{1}{c}{\textbf{Latency}} & \multicolumn{1}{c}{\textbf{Speedup}}\\
    
    \midrule
    \multirow{6}{*}{MBPP} & Target & 23.21 & 1.00x \\
    & dynHeur SL & 20.07 & 1.16x \\
    & static SL-5 & 15.88 & 1.46x \\
    & static SL-opt & 14.16 & 1.64x \\
    & \ourmethod (ours) & 12.58 & \bf{1.84x} \\
    \cmidrule(r){2-4}
    & oracle & 10.18 & {2.28x} \\

    \midrule
    \multirow{6}*{\parbox{1.5cm}{HumanEval (transfer learning)}} & Target & 23.46 & 1.00x \\
    & dynHeur SL & 22.57 &  1.04x \\
    & static SL-5 & 14.43 & 1.63x \\
    & static SL-opt & 14.42 & 1.63x \\
    & \ourmethod (ours) & 12.78 & \bf{1.84x} \\
    \cmidrule(r){2-4}
    & oracle & 10.59 & {2.22x} \\
    
    \midrule
    \multirow{6}*{CNN-DM} & Target & 38.29 & 1.00x \\
    & dynHeur SL & 21.18 & 1.81x \\
    & static SL-5 & 19.74 & 1.94x \\
    & static SL-opt & 20.66 & 1.85x \\
    & \ourmethod (ours) & 17.85 & \bf{2.15x} \\
    \cmidrule(r){2-4}
    & oracle & 15.41 & {2.48x} \\

    \midrule
    \multirow{6}*{Alpaca} & Target & 47.67 & 1.00x \\
    & dynHeur SL & 31.83 &  1.50x \\
    & static SL-5 & 23.79 & 2.00x \\
    & static SL-opt & 23.65 & 2.02x \\
    & \ourmethod (ours) & 22.49 & \bf{2.12x} \\
    \cmidrule(r){2-4}
    & oracle & 20.04 & {2.38x} \\

    \bottomrule[\heavyrulewidth] 
    \end{tabularx}
	\caption{Average latency results (in milliseconds) on different benchmarks. HumanEval results use a classifier trained on MBPP (transfer learning). All results are provided with greedy decoding (temperature=0).
 }
    \label{table:latency_results}
\vspace{-13mm}
\end{wraptable}
We also consider the optimal static SL baseline tuned on our validation sets (\textit{static SL-opt}). 
Finally, we also report results for our oracle (\cref{sec:spec_len}), which represents the lower bound on latency. 
\Cref{table:latency_results} presents our results using the rejection sampling scheme with greedy decoding (temperature=0) since baselines get higher speedup \citet{Leviathan2023}.
Employing an SL classifier consistently outperforms all other baselines across all benchmarks. Average latency improvements of \ourmethod over the optimal static SL and the dynamic heuristic baselines are 10.3\% and 31.4\% respectively, while preserving the same output as the target model. Importantly, our improvement does not come only from our training data: the optimal static SL (as fit by that data) is still underperformed by \ourmethod. Finally, \ourmethod transfers well across tasks: when trained on MBPP, it is still outperforms all baselines on HumanEval. See \cref{sec:analysis} for further analysis.





\section{Related Work}
Pioneering studies on speculative decoding~\cite{Leviathan2023,Chen2023_deep_mind} introduced a rejection sampling scheme that preserves the distribution of the target model, guaranteeing that speculative decoding maintains the quality of the target model.
Subsequent work~\cite{SpecInfer_Miao2023} elevated the average number of accepted tokens by using several draft models. 
Most recently, \citet{timor2024distributed} introduced DSI, a distributed variation of speculative decoding that is provably faster than non-distributed methods and does not require additional training or architectural changes. 
To eliminate the need for a separate draft model, \citet{Eagle_2024,Medusa_2024,speculative_streaming_2024,tiny_vicuna_2024} train additional, specialized draft layers on top of the transformer decoder. \ourmethod transfers well within domains and does not require classifier training per dataset whereas these methods necessitate training per dataset. At inference time, they employ a static SL; we believe that \ourmethod can be beneficially applied to these approaches, we leave this research for future work.
\citet{zhang2023draft} proposed draft-exiting with an adaptive threshold for self-speculative decoding using a rule-based approach that compares a confidence to a predetermined threshold. This method seems suitable to approaches where the draft is a subset of the target model, whereas our approach is more generic.
A very recent concurrent work by \citet{tandem2024} enhanced the draft model's accuracy by granting it access to the target model’s representations. In addition, it employed a classifier to determine whether to halt or continue the speculation process. Our work delves into the impact of the SL on the efficiency of speculative decoding, encompassing comparisons between static and dynamic SL approaches, as well as the upper bound of improvement represented by the oracle SL.

\section{Conclusion}
We have shown that using the same speculation lookahead parameter across speculative decoding iterations is suboptimal. 
We introduced \ourmethod, a dynamic speculation lookahead optimization method. The method uses a classifier that determines whether the draft model should continue to generate the next token or pause and transition to the target model for validation. We evaluated \ourmethod's effectiveness using four benchmarks and demonstrated average speedup gains of 10.3\% and 31.4\% relatively to the optimal static SL and dynamic heuristic baselines. Our results highlight the potential of further reducing inference cost by using simple, efficient techniques.






\bibliographystyle{plainnat}
\bibliography{custom}

\newpage
\appendix
\label{sec:appendix}
\section{Oracle SL Analysis}
\label{sec:appendix_oracle}

\begin{figure}
    \centering
    \begin{subfigure}{0.5\textwidth}
        \centering
        \includegraphics[width=\textwidth]{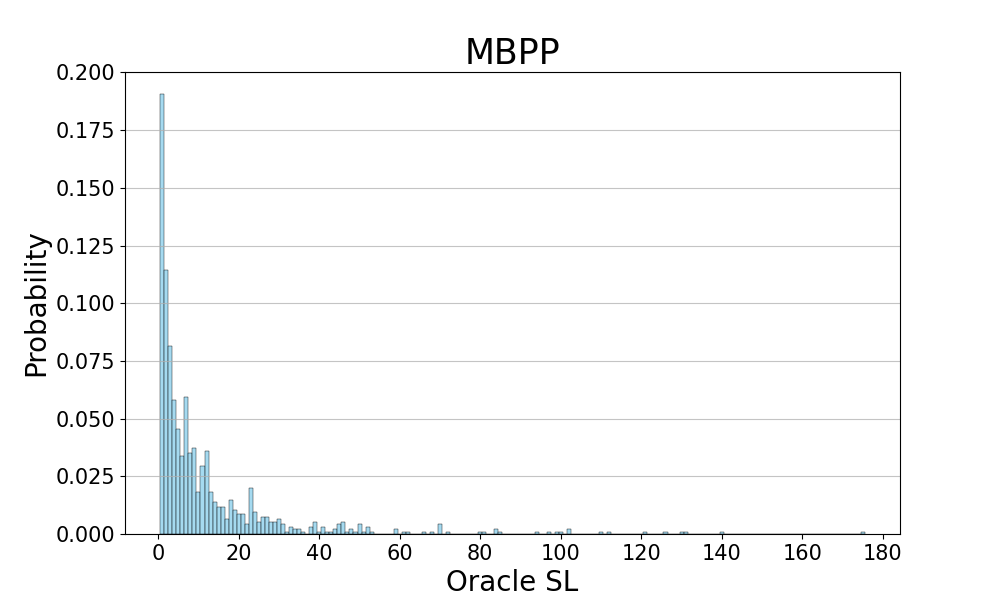}
    \end{subfigure}
    \hfill
    \begin{subfigure}{0.5\textwidth}
        \centering
       \includegraphics[width=\textwidth]{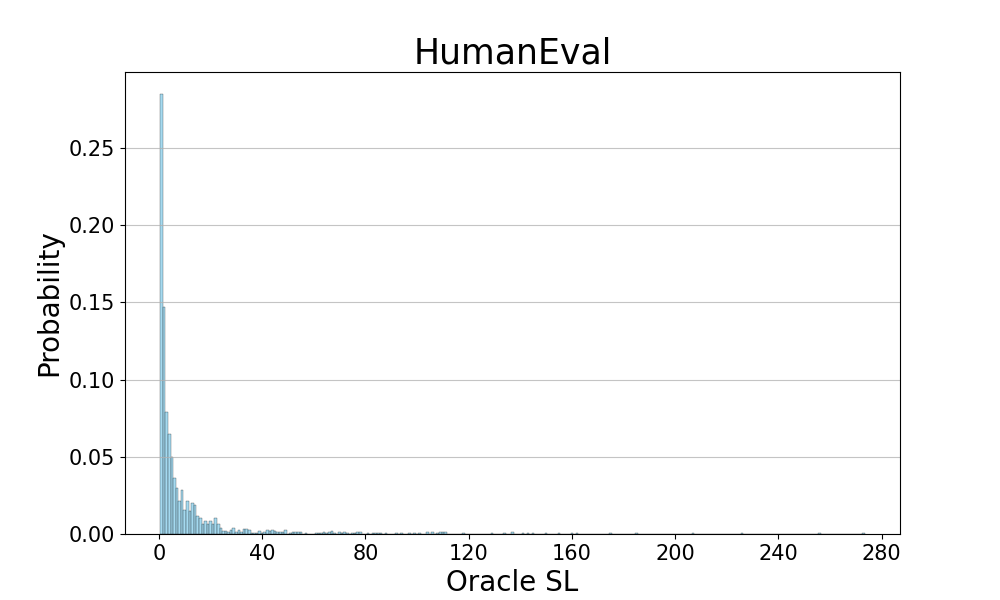}
    \end{subfigure}
    \hfill
    \begin{subfigure}{0.5\textwidth}
        \centering
        \includegraphics[width=\textwidth]{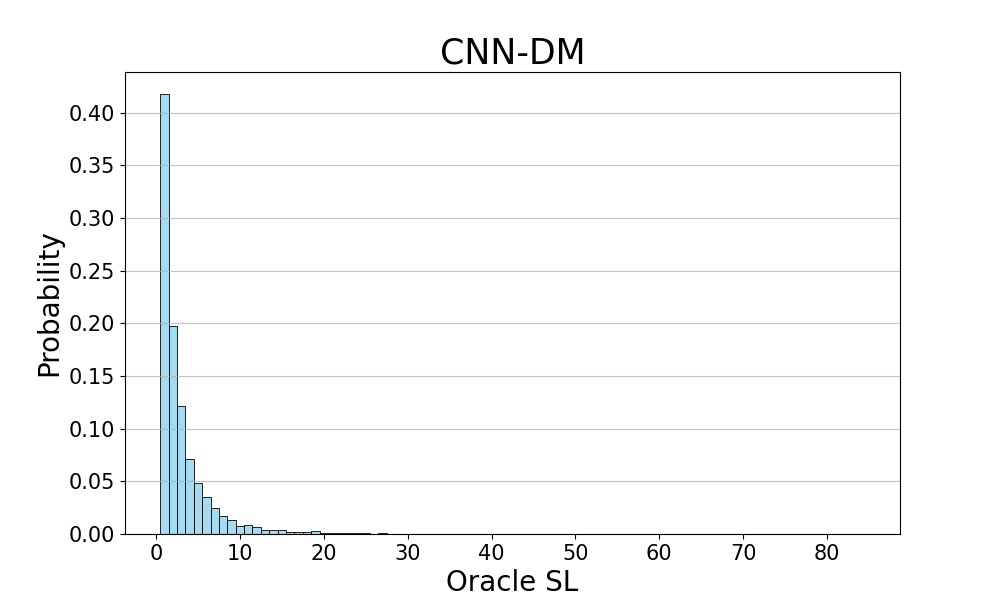}
    \end{subfigure}
    \hfill
    \begin{subfigure}{0.5\textwidth}
        \centering
        \includegraphics[width=\textwidth]{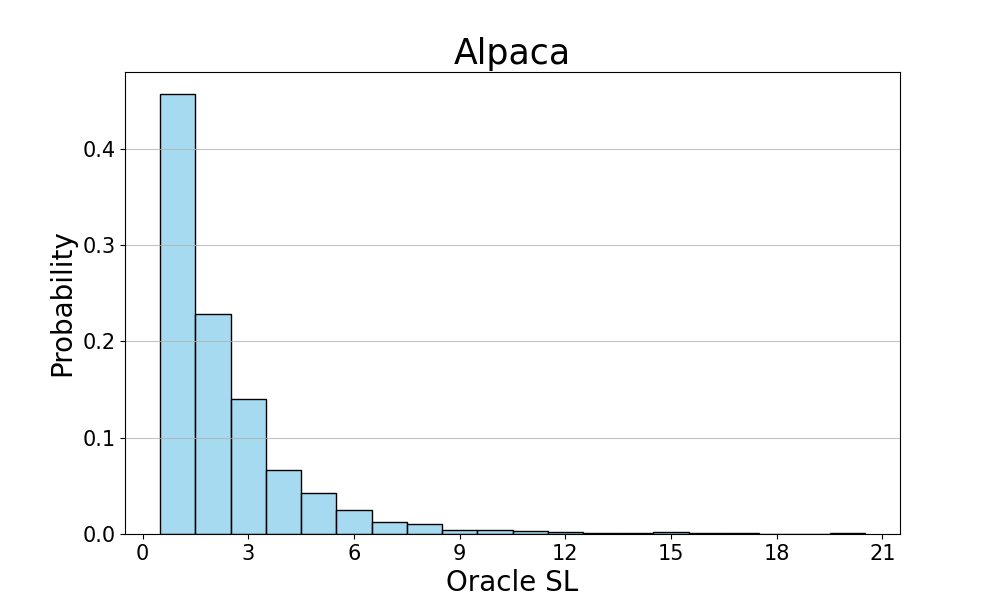}
    \end{subfigure}
    
    \caption{Oracle SL probability histogram on the different datasets. We observe a high variance of SL values.}
    \label{fig:oracle:mbpp}
\end{figure}

\Cref{tab:oracle} shows the oracle SL expectancy and standard deviation of the oracle measured on different datasets and models and \cref{fig:oracle:mbpp} shows the probability distribution of the oracle SL for the different datasets. We observe a high variance of SL values.

\begin{table}
    \centering
    \begin{adjustbox}{width=0.95\columnwidth,center}
    \begin{tabular}{c|c|c|c}
         \textbf{Datasets} & \textbf{Target Model}  & \textbf{Draft Model} & \textbf{ oracle SL} \\
         \hline
         MBPP & \starcodertarget & \starcoderdraft & 18.0 ($\pm$ 46.2) \\
         HumanEval & \starcodertarget & \starcoderdraft & 14.7 ($\pm$ 42.6) \\
         CNN-DM & \vicunatarget & \vicunadraft & \phantom{0}3.2 ($\pm$\phantom{0}4.0) \\
         Alpaca & \vicunatarget & \vicunadraft  & \phantom{0}2.4 ($\pm$\phantom{0}2.1) \\
    \end{tabular}
    \end{adjustbox}
    \caption{Average (and STD) of the oracle SL as determined by the oracle per dataset and target/draft models.}
    \label{tab:oracle}
\end{table}

\begin{figure}
    \centering
    \begin{subfigure}{0.95\columnwidth}
        \centering
        \includegraphics[width=\textwidth]{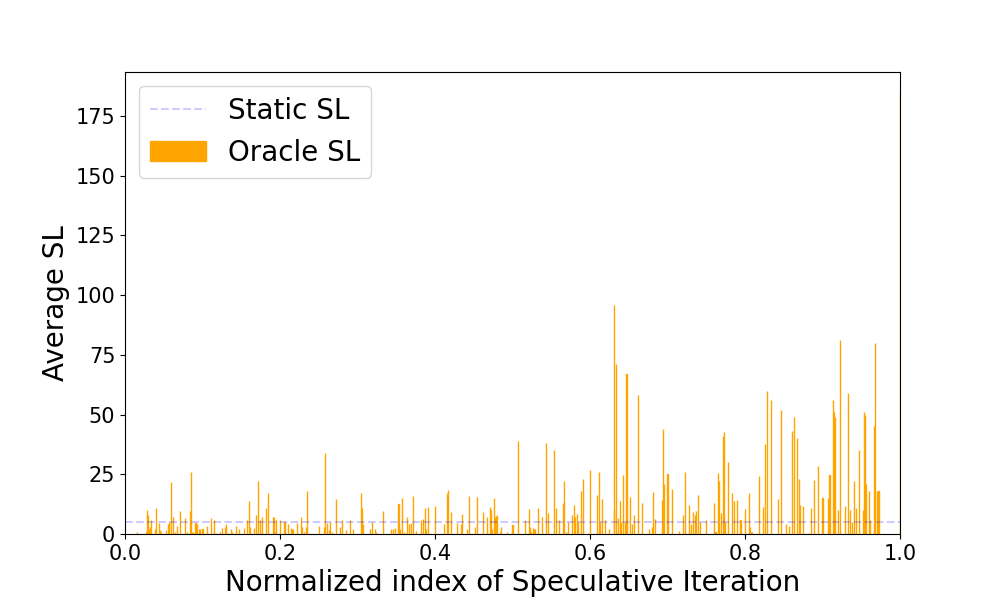}
        \subcaption{MBPP}
    \end{subfigure}
    \hfill
    \begin{subfigure}{0.95\columnwidth}
        \centering
       \includegraphics[width=\textwidth]{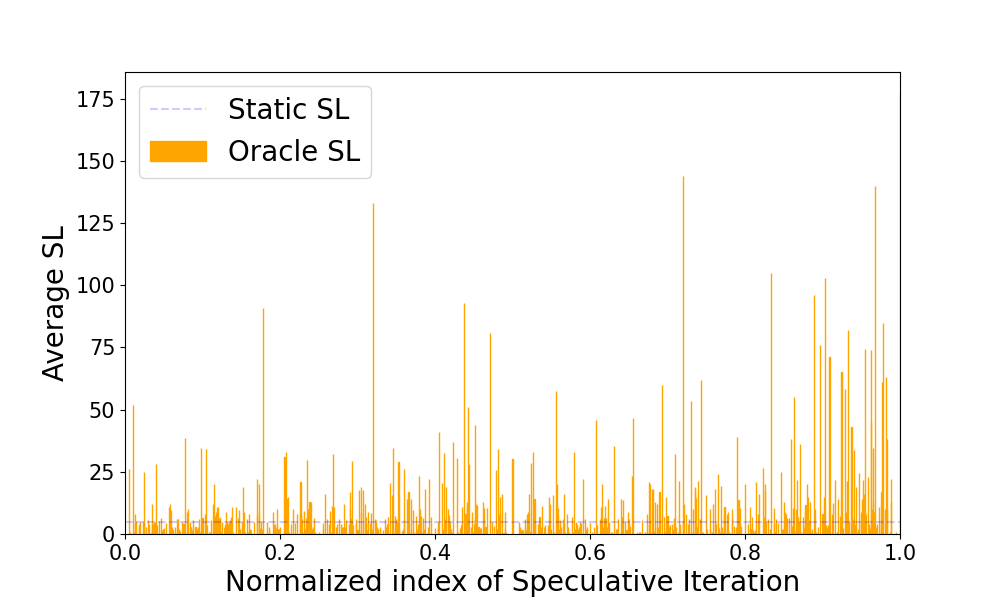}
        \subcaption{HumanEval}
    \end{subfigure}
    \hfill
    \begin{subfigure}{0.95\columnwidth}
        \centering
        \includegraphics[width=\textwidth]{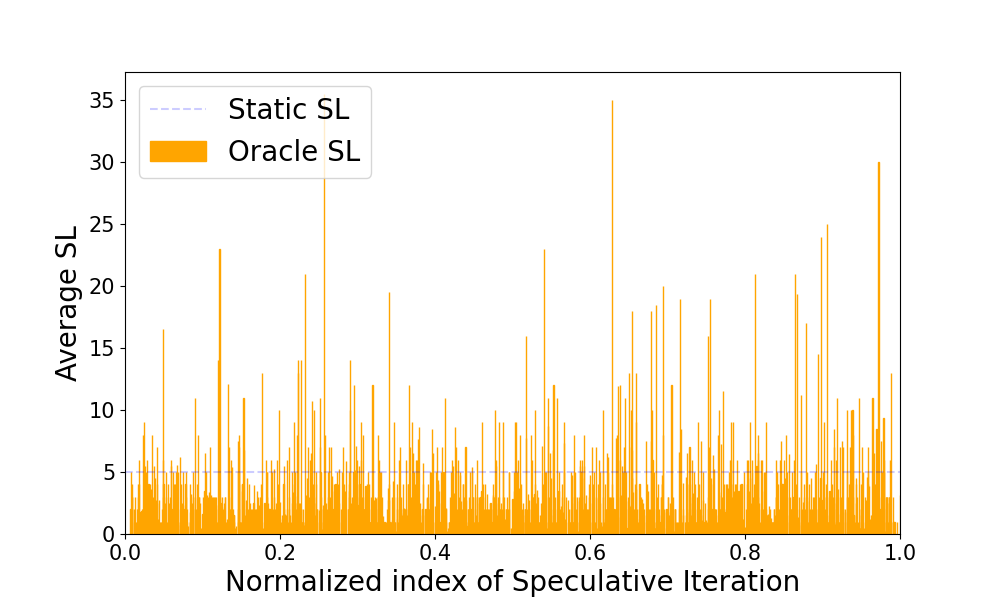}
        \subcaption{CNN-DM}
    \end{subfigure}
    \hfill
    \begin{subfigure}{0.95\columnwidth}
        \centering
        \includegraphics[width=\textwidth]{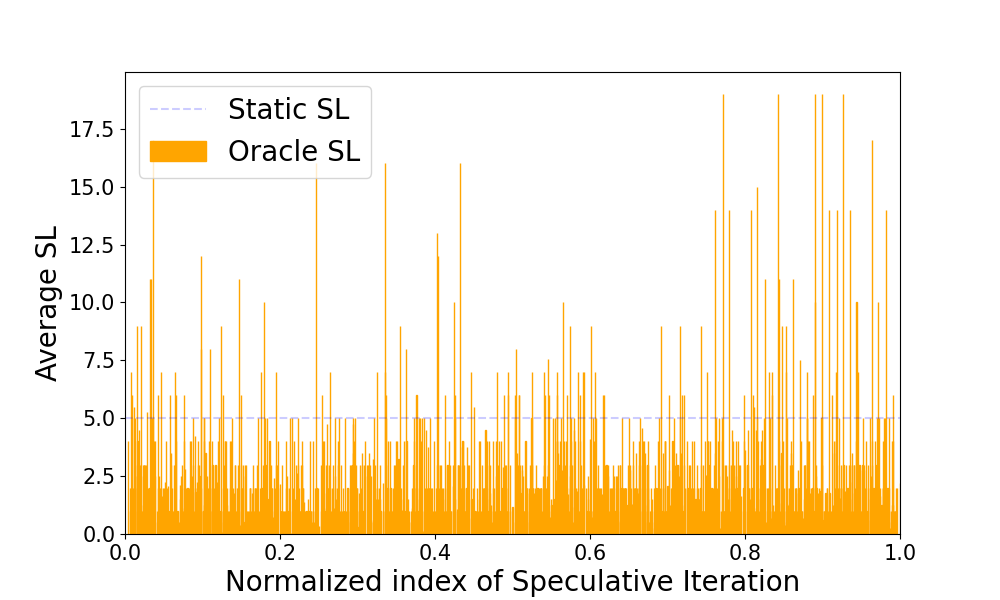}
        \subcaption{Alpaca}
    \end{subfigure}
    \caption{Bar chart of the average oracle SL. Note that the iteration index seems to have low predictive power for the oracle SL.
}
    \label{fig:oracle:avg}
\end{figure}

We hypothesized that later tokens are more predictive but eventually found only a relatively weak correlation, as \cref{fig:oracle:ex} and \cref{fig:oracle:avg} show. The figures are bar charts of the average oracle SL over the \textit{normalized index} of the Speculation Iteration.
We calculate the bars as follows.
For each prompt of a dataset, we have its corresponding sequence of oracle SLs.
The length of the sequence is equal to the number of Speculation Iterations.
For example, consider a prompt with oracle SLs $\langle 7, 3, 13, 21, 8 \rangle$.
Its normalized index is $\langle 0, 0.25, 0.5, 0.75, 1 \rangle$.
The bars represent the average oracle SL of buckets of size $0.0001$.

\section{Datasets and Prompts Details}
\label{sec:prompts}
We use standard datasets from \href{https://huggingface.co/datasets}{Hugging Face} and standard prompts from the state-of-of-the-art. 
\cref{tab:dataset} summarizes the composition of the datasets. We provide more details per dataset in the next sections.

\begin{table}[t!]
    \centering
    \small
    \begin{tabular}{c|ccc}
         \textbf{Datasets} & \textbf{Train}  & \textbf{Validation} & \textbf{Test} \\
         \hline
         MBPP & 374 & 80 & 80 \\
         HumanEval & - & - & 80 \\
         CNN-DM & 500 & 80 &  80 \\
         Alpaca & 500 & 80  &  80 \\
    \end{tabular}
    \caption{Number of samples per dataset and split.}
    \label{tab:dataset}
\end{table}

\subsection{MBPP}
For \href{https://huggingface.co/datasets/mbpp}{MBPP}, we use the `train', `validation' and `test' splits of the `full' subset. 
The whole `train' split is used for training, while 80 randomly selected samples of the `validation' and `test' splits are respectively used for validation and test. MBPP is distributed under the \href{https://creativecommons.org/licenses/by/4.0/deed.en}{cc-by-4.0 License}.

Concerning the prompt, we followed~\cite{bigcode-evaluation-harness, fried2023incoder} and included the description of the programming task and a single test to verify solution, in order to help the model catch the signature of the function (see \cref{fig:mbpp:prompt}).
\begin{figure}
\centering
\hrulefill
\begin{verbatim}
"""{text}
{test_list[0]}
"""
\end{verbatim}
\hrulefill
\caption{MBPP Prompt}
\label{fig:mbpp:prompt}
\end{figure}

\subsection{HumanEval}
\href{https://huggingface.co/datasets/openai_humaneval}{HumanEval}
dataset contains a single subset with a single split (`test' split). We use 80 randomly selected samples of that split for test. Note that since we evaluate transfer learning from MBPP, we don't need HumanEval training and validation sets. HumanEval is distributed under the \href{https://choosealicense.com/licenses/mit/}{MIT License}.

Prompt contains only \texttt{prompt} field from the dataset.

\subsection{CNN-DM}
For \href{https://huggingface.co/datasets/cnn_dailymail}{CNN-DM}, we use the `train', `validation' and `test'  splits of the `2.0.0' subset. 
500 randomly selected samples of the `train' split is used for training, while 80 randomly selected samples of the `validation' and `test' splits are respectively used for validation and test. CNN-DM is distributed under the \href{https://choosealicense.com/licenses/apache-2.0/}{Apache License 2.0}.

We included the \texttt{article} field in the prompt as in \cref{fig:cnndm:prompt}. 
\begin{figure}
\centering
\hrulefill
\begin{verbatim}
"""Summarize:
{article}
Summary:
"""
\end{verbatim}
\hrulefill
\caption{CNN-DM Prompt}
\label{fig:cnndm:prompt}
\end{figure}

\subsection{Alpaca}
\href{https://huggingface.co/datasets/tatsu-lab/alpaca}{Alpaca} dataset contains a single split (`train' split). As for CNN-DM, 500 randomly selected samples of the `train' split is used for training, while 80 randomly selected samples of the `validation' and `test' splits are respectively used for validation and test. Alpaca is distributed under the \href{https://creativecommons.org/licenses/by-nc/4.0/deed.en}{cc-by-nc-4.0 License}.

We follow~\citet{alpaca2023} to define the prompts. For samples with a non-empty input field, we use the prompt as in \cref{fig:alpaca1:prompt} while for samples with empty input field, we use the prompt as in \cref{fig:alapca2:prompt}.
\begin{figure}
\centering
\hrulefill
\begin{verbatim}
 """Below is an instruction that describes a
 task, paired with an input that provides
 further context. Write a response that
 appropriately completes the request.

### Instruction:
{instruction}

### Input:
{input}

### Response:
"""

\end{verbatim}
\hrulefill
\caption{Alpaca prompt for samples with a non-empty input field.}
\label{fig:alpaca1:prompt}
\end{figure}

\begin{figure}
\centering
\hrulefill
\begin{verbatim}
"""Below is an instruction that describes a
task. Write a response that appropriately
completes the request.

### Instruction:
{instruction}

### Response:
"""
\end{verbatim}
\hrulefill
\caption{Alpaca prompt for samples with empty input field.}
\label{fig:alapca2:prompt}
\end{figure}

\section{Models}
\label{sec:models}

For all models, we retrieve model weights from \href{https://huggingface.co/models}{Hugging Face}.
For clarity and reproducibility, we provide the URLs for each model used:\\
\begin{itemize}
\item \vicunatarget:  \url{https://huggingface.co/lmsys/vicuna-13b-v1.3}, distributed under Non-Commercial License.  
\item \vicunadraft: \url{https://huggingface.co/double7/vicuna-68m}, distributed under the \href{https://choosealicense.com/licenses/apache-2.0/}{Apache License 2.0}. 
\item \starcodertarget: \url{https://huggingface.co/bigcode/starcoder}, distributed under the \href{https://www.licenses.ai/}{Responsible AI License}.
\item \starcoderdraft: \url{https://huggingface.co/bigcode/tiny_starcoder_py}, also distributed under the \href{https://www.licenses.ai/}{Responsible AI License}
\end{itemize}

\section{Classifier}
\label{sec:app_cls}

\subsection{Feature Extraction from Training Data for Classifier Training}
To train the model for each dataset, MBPP, CNN-DM, and Alpaca, we used the corresponding training set of each dataset. For each token in each training set, we extracted a boolean label (accepted/rejected) and a list of features in the following manner: 
We ran the target model using the standard autoregressive approach for generating tokens based on the input prompt. In contrast, the draft model iteratively generated only a single token per iteration. During each iteration, the draft model generated this token based on the concatenation of the input prompt and the tokens subsequently generated by the target model. Draft tokens that resembled the target token at the same position were labeled "accepted," while others were labeled "rejected."
Corresponding features were extracted for both draft and target tokens, encompassing the top-k probabilities of the vocabulary distribution, the entropy associated with these probabilities, and the token position value counted from the beginning of the generation process.

\subsection{Classifier F1 Results}

We evaluate the quality of the classifier by measuring its F1 score on the validation set and report in \cref{tab:F1} F1 scores for both static SL and \ourmethod. F1 scores of \ourmethod always outperforms F1 scores of static SL.

F1 score measures the accuracy of the classifier in predicting the speculative length (SL) but does not account for how well the predictions align with the oracle’s behavior in reducing latency. In particular, F1 is influenced by 2 different types of error FP and FN that have a different impact on the speedup. FP Errors lead to unnecessary speculative execution, which wastes resources but might not drastically reduce overall speedup. FN Errors lead to missed opportunities for speedup, having a more severe impact on overall latency reduction.

\begin{table}
    \centering
    \small
    \begin{tabularx}{0.5\linewidth}{ccc}
    \toprule[\heavyrulewidth]\toprule[\heavyrulewidth]
         \textbf{Datasets} & \textbf{F1 static-SL}  & \textbf{F1 \ourmethod} \\
         \hline
         MBPP & 85 & 95 \\
         CNN-DM & 76 & 88 \\
         Alpaca & 68 & 77 \\
         \bottomrule[\heavyrulewidth] 
    \end{tabularx}
    \caption{Classifier F1 scores for both static SL and \ourmethod. }
    \label{tab:F1}
\end{table}

\section{Additional Implementation Details}
\label{sec:app_impl}
Our implementation is based on the \href{https://github.com/huggingface/transformers}{Transformers library of HuggingFace}, distributed under the \href{https://choosealicense.com/licenses/apache-2.0/}{Apache License 2.0}, and \href{https://github.com/pytorch/pytorch}{PyTorch Deep Learning library}, distributed under the \href{https://opensource.org/license/BSD-3-Clause}{BSD License (BSD-3)}.
Our code will be available upon publication under the \href{https://choosealicense.com/licenses/apache-2.0/}{Apache License 2.0}.

For every dataset, \ourmethod classifier is trained on the train set; threshold $\tau$ and $\text{SL}_{\max}$ hyper-parameters are fine-tuned on the validation set optimizing the latency. 
Optimal static SL is estimated on the validation set. Latency results are reported on the test set.

All our  experiments are run on a single A100 80GB GPU. 

\section{Additional Results}
\label{sec:app_adtnl_results}

Table \ref{table:results_percentage} shows the percentage of improvement in latency of \ourmethod over static SL-opt and dynHeur SL baselines. The numbers are calculated based on the latency results shown in Table \ref{table:latency_results}

\begin{table}[t!]
	\centering
	\small
    \begin{tabularx}{0.5\linewidth}{lcc} 
    \toprule[\heavyrulewidth]\toprule[\heavyrulewidth]

    \textbf{Dataset} & \textbf{static SL-opt} & \textbf{dynHeur SL}
    \\
    \midrule
   
    MBPP & 11.2 & 37.3\\
    HumanEval & 11.4 & 43.4 \\
    CNN-DM  & 13.6 & 15.7\\
    Alpaca &  4.9 & 29.3 \\
   \midrule
    Average & \textbf{10.3} & \textbf{31.4} \\
    
    \bottomrule[\heavyrulewidth] 
    \end{tabularx}
    \caption{The latency improvement (percentage) of \ourmethod over static SL-opt and dynHeur SL across four datasets.
    }
    \label{table:results_percentage}
\end{table}

\section{Further Latency Results Analysis}
\label{sec:analysis}

Concerning SL-opt and SL-5 speedup values on CNN-DM reported in \cref{table:latency_results}, note that the optimal static SL is tuned on the validation set while the latency and speedup reported numbers are on the test set. This is similar to our setup, where the classifier is tuned on the validation set, and similarly reported on the test set. We observe that for CNN-DM, the optimal SL on the validation set is not as good as the SL-5 on the test set.

Since \ourmethod is sampling temperature agnostic, we present in \cref{table:latency_analysis} latency results for CNN-DM with non-zero temperature; we observe that \ourmethod speedup also improves in that case. In addition, we report latency results using a simple rule-based approach: we use the perplexity to measure the confidence in the sequence of tokens predicted by the draft, when lower perplexity indicates higher confidence. Optimal perplexity threshold is tuned on our validation sets (\textit{ppl SL-opt}). We observe that it yields a lower speedup. 


    
	

\begin{table}[t!]
    \centering
    \small
    \begin{tabularx}{0.6\linewidth}{p{1.5cm}lcc} 
    \toprule[\heavyrulewidth]\toprule[\heavyrulewidth]

    \textbf{Benchmark} & \textbf{Method} & \multicolumn{1}{c}{\textbf{Latency(ms)}} & \multicolumn{1}{c}{\textbf{Speedup}}\\
    \midrule
    \multirow{5}*{CNN-DM} & Target & 36.32 & 1.00x \\
    & static SL-5  & 21.88 & 1.66x \\
    & \ourmethod (ours) & 19.96 & \bf{1.82x} \\
    \cmidrule(r){2-4}
    & ppl SL-opt & 26.79 & 1.43x \\
    
    \bottomrule[\heavyrulewidth] 
    \end{tabularx}
	\caption{Additional average latency results for CNN-DM: temperature=1; perplexity-based baseline.
 }
    \label{table:latency_analysis}
\end{table}

\end{document}